\documentclass{bmvc2k}

\title{UMAD: Unsupervised Mask-Level\\Anomaly Detection for Autonomous Driving}

\addauthor{Daniel Bogdoll\textsuperscript{\textasteriskcentered}}{bogdoll@fzi.de}{1,2}
\addauthor{Noël Ollick\textsuperscript{\textasteriskcentered}}{noel.ollick@student.kit.edu}{1}
\addauthor{Tim Joseph}{joseph@fzi.de}{2}
\addauthor{Svetlana Pavlitska}{pavlitska@fzi.de}{1,2}
\addauthor{J. Marius Zöllner}{zoellner@fzi.de}{1,2}

\addinstitution{
Karlsruhe Institute of Technology (KIT)\\
Karlsruhe, Germany
}
\addinstitution{
FZI Research Center for Information Technology\\
Karlsruhe, Germany
}

\runninghead{D. Bogdoll et al.}{UMAD: Unsupervised Mask-Level Anomaly Detection}

\usepackage{graphicx}
\usepackage{wrapfig}
\usepackage{booktabs}
\usepackage{amssymb}
\usepackage{booktabs}

\begin{document}

\maketitle

\def\thefootnote{\textsuperscript{\textasteriskcentered}}\footnotetext{These authors contributed equally}\def\thefootnote{\arabic{footnote}}

\begin{abstract}
Dealing with atypical traffic scenarios remains a challenging task in autonomous driving. However, most anomaly detection approaches cannot be trained on raw sensor data but require exposure to outlier data and powerful semantic segmentation models trained in a supervised fashion. This limits the representation of normality to labeled data, which does not scale well. In this work, we revisit unsupervised anomaly detection and present UMAD, leveraging generative world models and unsupervised image segmentation. Our method outperforms state-of-the-art unsupervised anomaly detection.
\end{abstract}

\section{Introduction}
\label{sec:intro}

Although great achievements have been made in autonomous driving, reacting to the unknown remains a significant challenge~\cite{Heidecker_2021_An_Application_IV, Bogdoll_2022}. \citet{Heidecker_2021_An_Application_IV} categorize anomalies into the sensor, content, and temporal layer: Anomalies in the sensor layer are related to sensory abnormalities, anomalies in the content layer regard abnormalities in single observations, such as atypical objects, and the temporal layer considers behavioral anomalies in the context of multiple frames.

Classically, anomaly detection is based on highly specialized methods, focusing on the content layer~\cite{Nayal_2023_RbA_ICCV, delicOutlierDetectionEnsembling2024,Nekrasov_2023_UGainS_GCPR}. However, a perpendicular line of work tries to learn a more general understanding of the world. Generative world models have shown promising results in autonomous driving~\cite{Hu_2023_Neural_Phd, Hu_2023_gaia1_arXiv, Bogdoll_2023_MUVO_arXiv, Zhang_2023_Copilot4d_ICLR, gaoVistaGeneralizableDriving2024}. They embed sensory data into latent states, reconstruct observations based on those, and predict action-conditioned future states. For anomaly detection, however, they have not been utilized yet~\cite{Bogdoll_2023_Exploring_SSCI}. In this paper, we address to which extent world models and unsupervised image segmentation can be used for anomaly detection in autonomous driving and, contrary to many prior anomaly detection models in this domain~\cite{Blum_2019_Fishyscapes_IJCV, DiBiase_2021_Pixelwise_CVPR, Grcic_2022_DenseHybrid_ECCV}, propose an unsupervised anomaly detection method which does not rely on outliers in the training data. We present \textbf{U}nsupervised \textbf{M}ask-Level \textbf{A}nomaly \textbf{D}etection for Autonomous Driving~(UMAD), leveraging generative world models and segmentation models. In the experimental setup for this paper, UMAD utilizes the multimodal world model MUVO \cite{Bogdoll_2023_MUVO_arXiv}, which was trained on data from the CARLA simulator \cite{Dosovitskiy17}. For refined localization of anomalies, UMAD leverages masks which are generated with the unsupervised image segmentation approach U2Seg \cite{Niu_2023_Unsupervised_arXiv}. We also provide experimental results for the zero-shot segmentation model SAM \cite{Kirillov_2023_segany_arXiv}. UMAD improves upon the current SotA by achieving an FPR$_{95}$ reduction of $36.90 \%$ on the AnoVox benchmark, setting a new baseline in unsupervised anomaly detection for autonomous driving.


\section{Related Work}
\label{sec:sota}

Recent trends in anomaly detection have shown that utilizing semantic segmentation models and including Out-of-Distribution (OOD) data during training achieves close-to-perfect benchmark results~\cite{Blum_2019_Fishyscapes_IJCV,Chan_2021_SegmentMeIfYouCan_NIPSTDB,delicOutlierDetectionEnsembling2024}. However, we argue that normality should be learned on raw sensory data and thus in an unsupervised fashion. Including anomalies during training poses the risk of missing anomalies in a never-ending open-world setting, and utilizing supervised semantic segmentation~\cite{Nayal_2023_RbA_ICCV,ackermannMaskomalyZeroShotMask2023a,DiBiase_2021_Pixelwise_CVPR}, bounding boxes~\cite{liuHybridVideoAnomaly2021,fangTrafficAccidentDetection2022}, or language~\cite{tian_road} limits the definition of normality to labeled training data, which does not scale well. Here, we revisit the field of unsupervised anomaly detection and explore mask-level approaches.

\textbf{Unsupervised Anomaly Detection.} While modeling uncertainty of models on computer vision tasks in an unsupervised way has already been addressed~\cite{Kendall_2017_What_NIPS, Gal_2016_Dropout_ICML, Gal_2017_Concrete_NIPS, Gustafsson_2020_Evaluating_CVPR_Workshops}, these models were not evaluated on anomaly detection benchmarks, but regarding their eligibility to model uncertainty of general computer vision tasks. Since anomaly detection is not only relevant in autonomous driving, there are also unsupervised anomaly detection methods in other domains. For example, \citet{Zhou_2020_Encoding_ECCV} have developed an anomaly detection model on retinal images, e.g., for detecting retinal diseases or lesions, and \citet{Wang_2020_Image_ApplSc} have evaluated their anomaly detection model on the MVTec AD dataset~\cite{Bergmann_2019_MVTec_IEEECVF} for industrial inspection. Similarly, self-supervised detection methods exist in such a setting~\cite{schwartz2024maeday,jiang2022masked,zavrtanik2021reconstruction}. Others used the MNIST~\cite{LeCun_1998_Gradient_based_IEEE,An_2015_Variational_Special_Lecture_IE, Hendrycks_2017_Baseline_ICLR} or CIFAR~\cite{Krizhevsky_2009_Learning,Hendrycks_2017_Baseline_ICLR,Vu_2019_Anomaly_arXiv} datasets which contain images of only small sizes for their evaluation.

In anomaly detection in the surveillance setting, there is also a trend towards supervision requiring labeled training data~\cite{liuHybridVideoAnomaly2021,fangTrafficAccidentDetection2022}. However, there are two recent unsupervised methods. \citet{Abati_2019_Latent_IEEECVF} have developed a novelty detection model that uses a deep autoencoder in combination with an autoregressive parametric density estimator, using real world data with the UCSD Ped2~\cite{Chan_2008_Modeling_IEEETPAMI} and the ShanghaiTech~\cite{Luo_2017_Revisit_ICCV} datasets. Similar to \citet{Abati_2019_Latent_IEEECVF}, \citet{Park_2020_Learning_IEEECVF} trained MNAD on datasets with images from the real world \cite{Chan_2008_Modeling_IEEETPAMI, Luo_2017_Revisit_ICCV, Lu_2013_Abnormal_IEEEICCV}, which partly contain semantic classes that can also be found in autonomous driving, e.g., pedestrians, bicycles, and cars. They compare the reconstruction of an autoencoder to the initial input image by using the L2 distance and the peak signal-to-noise ratio~(PSNR) in order to calculate abnormality scores. 

\textbf{Mask-Level Anomaly Detection.} A trend to improve anomaly detection methods is to use learned masks to generate instance-level detections. For detecting masks of anomalous instances in an image, the zero-shot Segment Anything Model~(SAM) by \citet{Kirillov_2023_segany_arXiv} was quickly used for the localization of anomalies in images. Here, we give an overview of recent methods using segmentations during post-processing, as shown in Table~\ref{tab:mask_sota}.

\begin{table}[h]
\centering
\resizebox{0.9\textwidth}{!}{%
\begin{tabular}{@{}lccccl@{}}
\toprule
\textbf{Method} & \textbf{Supervision} & \textbf{Temporal} & \textbf{Multimodal} & \textbf{OOD Data} & \textbf{Extra Networks} \\ \midrule
SAA+~\cite{caoSegmentAnyAnomaly2023}            & \checkmark  & ---  & \checkmark                 & ---               &       \checkmark~\cite{liu2023grounding,Kirillov_2023_segany_arXiv}                \\
UGains~\cite{Nekrasov_2023_UGainS_GCPR}          & \checkmark & ---  & ---                  & ---              & \checkmark~\cite{Nayal_2023_RbA_ICCV,Kirillov_2023_segany_arXiv}                      \\
S2M~\cite{Zhao_2023_segment_arXiv}             & \checkmark & ---  & ---                  & \checkmark               & \checkmark~\cite{renFasterRCNNRealTime2015,liuResidualPatternLearning2023a,Kirillov_2023_segany_arXiv}                     \\
ClipSAM~\cite{liClipSAMCLIPSAM2024}         & \checkmark  & --- & \checkmark                  & ---               & \checkmark~\cite{radfordLearningTransferableVisual2021,Kirillov_2023_segany_arXiv}                       \\ \midrule
\textbf{UMAD}          & ---  & \checkmark  & \checkmark               & ---               & \checkmark~\cite{Bogdoll_2023_MUVO_arXiv,Niu_2023_Unsupervised_arXiv}                      \\ \bottomrule
\vspace{2pt}
\end{tabular}%
}
\caption{\textbf{Overview of mask-level anomaly detection methods}. The table shows methods that use segmentation masks for post-processing. Supervision refers to the necessity of labeled data during training. Temporality denotes the ability of a method to incorporate temporal context. Multimodal models utilize further modalities, such as text or lidar data, for anomaly detection. OOD data shows whether outliers were needed during training. Finally, all external networks are shown.}
\label{tab:mask_sota}
\end{table}

Segment Any Anomaly~(SAA+)~\cite{caoSegmentAnyAnomaly2023} utilizes pre-trained foundation models for mask-level anomaly detection without further training. They first employ Grounding DINO~\cite{liu2023grounding}, which provides bounding boxes for regions defined by a prompt. To refine those box regions into masks, they utilize SAM~\cite{Kirillov_2023_segany_arXiv}. Similarly, S2M~\cite{Zhao_2023_segment_arXiv} proposes bounding boxes that include anomalies, followed by SAM. Similar to many other anomaly detection models, they use outlier exposure during training. UGainS~\cite{Nekrasov_2023_UGainS_GCPR} uses the existing anomaly detection model Rejected by All (RbA)~\cite{Nayal_2023_RbA_ICCV} in combination with SAM for localizing anomalous instances in the observation. Finally, ClipSAM~\cite{liClipSAMCLIPSAM2024} utilizes CLIP text and image encoders~\cite{radfordLearningTransferableVisual2021} to generate an initial anomaly mask and refine it with SAM.

\textbf{Research Gap.} In autonomous driving, recent trends have moved away from unsupervised anomaly detection~\cite{delicOutlierDetectionEnsembling2024, DiBiase_2021_Pixelwise_CVPR, Grcic_2022_DenseHybrid_ECCV}, and benchmarks are saturated with near-perfect results~\cite{Blum_2019_Fishyscapes_IJCV,Chan_2021_SegmentMeIfYouCan_NIPSTDB,delicOutlierDetectionEnsembling2024}. While unsupervised anomaly detection methods from other domain methods are available, to the best of our knowledge, there exists no unsupervised anomaly detection model for autonomous driving. Similarly, the recent trend of mask-level anomaly detection methods works in a supervised manner. Thus, we see a clear need to revisit the field of unsupervised anomaly detection in order to use vast amounts of unlabeled data for training, as typically available in autonomous driving.

\section{Method}
\label{sec:method}

\begin{figure}[h!]
        \centering
        \includegraphics[width=\textwidth]{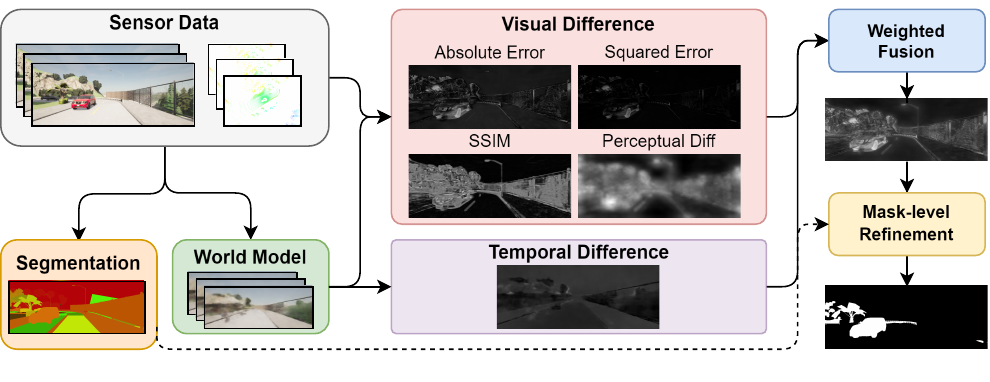}
        \caption{\textbf{Overview of UMAD.} First, multimodal sensor data is fed into a world model to reconstruct and predict frames, and semantic masks are derived from camera data. \textbf{Visual Differences} are used to compare a reconstruction of the current observation to the accompanying sensor data frame. The \textbf{Temporal Difference}, on the contrary, solely compares multiple future predictions from the world model. The pixel-wise scores are then fused and the resulting anomaly map is refined based on the generated masks.}
        \label{fig:overview}
    \end{figure}

As we have shown in Section~\ref{sec:sota} and Table~\ref{tab:mask_sota}, UMAD is the first unsupervised mask-level anomaly detection method in the context of autonomous driving, which means that UMAD can be trained purely based on unlabeled sensor recordings, that do not have to contain abnormal driving situations. As shown in Figure~\ref{fig:overview}, first, UMAD takes multimodal data from several different sensors such as a camera and lidar sensor as input for a world model to reconstruct and predict future frames. Furthermore, semantic masks are derived from camera data. More details on the encoder-decoder architecture of the utilized world model MUVO can be found in~\cite{Bogdoll_2023_MUVO_arXiv}.

For \textit{visual differences}, a reconstruction of the current observation is compared to the accompanying sensor data frame based on multiple methodologies. For \textit{temporal differences}, only multiple future predictions from the world model are compared. After a weighted fusion of the pixel-wise scores, the resulting anomaly map is refined based on the generated masks.

UMAD first uses the world model to generate a reconstruction of the current frame. This reconstruction is then compared to the ground truth sensory data from the camera sensor of the autonomous vehicle. While UMAD only uses camera data, the world model is grounded and conditioned by further sensor modalities, planned actions, and the provided route. To compute \textit{visual differences}, we employ several image comparison methods. The absolute error~$\Delta_{ABS}$ and mean squared error~$\Delta_{MSE}$ are calculated for each pixel individually and measure the differences in the $r,g,b$ color channels of the reconstruction and the sensory image. In Eq.~\ref{eq:abs} and~\ref{eq:mse}, $\tilde{z}_x$ denotes a ground truth value and $\hat{z}_{\hat{x}}$ a predicted value.

\begin{equation}
    \Delta_{ABS} = \frac{|\tilde{r}_x - \hat{r}_{\hat{x}}| + |\tilde{g}_x - \hat{g}_{\hat{x}}| + |\tilde{b}_x - \hat{b}_{\hat{x}}|}{3}
    \label{eq:abs}
\end{equation}

\begin{equation}
\Delta_{MSE} = \frac{(\tilde{r}_x - \hat{r}_{\hat{x}})^2 + (\tilde{g}_x - \hat{g}_{\hat{x}})^2 + (\tilde{b}_x - \hat{b}_{\hat{x}})^2}{3}
    \label{eq:mse}
\end{equation}

Contrary to this, the difference based on the Structural Similarity Index~$\Delta_{SSIM}$~\cite{Wang_2004_ssim_IEEE} compares two images based on their structure by utilizing batches of multiple proximate pixels. We compare sliding window patches between the ground truth $x$ and the prediction $\hat{x}$. In Eq.~\ref{eq:ssim}, $\mu$ denotes means and $\sigma$ (co)variances. The constants $\kappa_1$ and $\kappa_2$ are added for numerical stability~\cite{vojir2021, Wang_2004_ssim_IEEE}.

\begin{equation}
\Delta_{SSIM} = \frac{(2\mu_x\mu_{\hat{x}} + \kappa_1)(2\sigma_{x \hat{x}} + \kappa_2)}{(\mu_x^2 + \mu_{\hat{x}}^2 + \kappa_1)(\sigma_x^2 + \sigma_{\hat{x}}^2 + \kappa_2)}
\label{eq:ssim}
\end{equation}

Finally, perceptual difference $\Delta_{PD}$~\cite{DiBiase_2021_Pixelwise_CVPR} is an image comparison method that leverages a pre-trained deep convolutional network to extract features and compare two images pixel-wise based on their content. Similar to \citet{DiBiase_2021_Pixelwise_CVPR}, we utilize weights which are pre-trained on the ImageNet \cite{deng_imagenet} dataset. In Eq.~\ref{eq:pd}, $F^i$ denotes the $i$-th layer of a VGG network, and $M$ and $N$ refer to elements and layers, respectively. 

\begin{equation}
\Delta_{PD} = \sum_{i = 1}^{N} \frac{1}{M_i} \lVert F^{i}(x) - F^{i}(\hat{x}) \rVert_1
    \label{eq:pd}
\end{equation}

For \textit{temporal differences}~$\Delta_{TD}$, we compare multiple predictions of the world model to each other. The temporal difference is calculated by comparing prior predictions for the current time step to each other. For this, the mean of the absolute errors between $n$ past predictions $\hat{z}_{t-i}$ for time $t$ and the current reconstruction $\hat{z}_t$ is computed, as shown in Eq.~\ref{eq:td}. 

\begin{equation}
\Delta_{TD} =
\frac{1}{n}\left(\sum_{i=1}^{n}
\Delta_{ABS}\left(\hat{z}_{t-i},\hat{z}_t\right)\right)
\label{eq:td}
\end{equation}

All difference maps are then normalized and can be fused by assigning weights~$ w_i \in [0,1]$ to each of them. While the resulting anomaly map assigns anomaly scores to each pixel in the image, it does not classify instances in an observation as anomalous. For this, we refine the scores with instance masks to generate mask-level anomaly scores. By utilizing an image segmentation approach for mask generation, UMAD iterates through each predicted mask and calculates average instance-wise anomaly scores. More details can be found in~\cite{ollick_camera_2024_BA}.

\section{Experiments}
\label{sec:exp}

While there are common anomaly benchmarks in the context of autonomous driving, such as Fishyscapes~\cite{Blum_2019_Fishyscapes_IJCV} or SegmentMeIfYouCan~\cite{Chan_2021_SegmentMeIfYouCan_NIPSTDB}, they are limited to camera data and do not contain data on actions, e.g., steering wheel angle, or additional sensory data. Among existing anomaly detection benchmarks~\cite{bogdoll_percdata}, the recent AnoVox anomaly detection benchmark~\cite{bogdoll2024anovox} is the only benchmark containing multimodal sensory data and action data of the ego-vehicle.

\begin{figure}[h]
    \centering
    \includegraphics[width=1\textwidth]{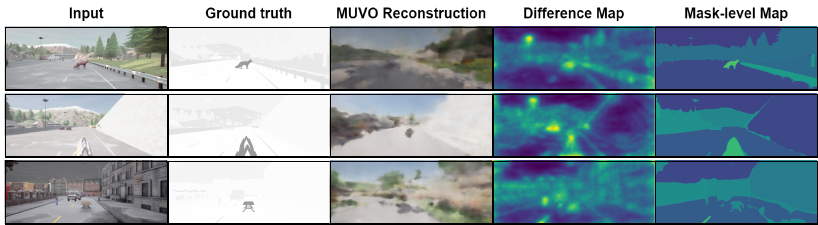}
    \caption{\textbf{Exemplary Detections}. The first columns show the input image and the corresponding ground truth. MUVO reconstructions are utilized to generate difference maps, which are finally refined to mask-level maps. Masks are generated by the unsupervised segmentation model U2Seg. The first two rows show positive cases, while the third row shows a failure case.}
    \label{fig:eval_examples}
\end{figure}

\textbf{Benchmark.} The AnoVox benchmark~\cite{bogdoll2024anovox} contains both static and temporal behavioral anomalies. Here, we only generate a subset with static anomalies, i.e., unexpected objects on the road. It includes images, lidar point clouds, routemaps, panoptic segmentation maps and was created using the CARLA simulator. For evaluation, we generated 16 abnormal driving scenarios with 200 frames each using the provided framework for generating a small-size dataset with anomalies comparable to current benchmarks. The scenarios take place in different towns under different weather conditions and contain static anomalies, e.g., an object or an animal standing on the street, as depicted in Figure~\ref{fig:eval_examples}. The dataloader for the world model samples each $10^{th}$ frame, i.e., every second. 

\textbf{Experimental Setup.} UMAD requires both an unsupervised world model and an unsupervised segmentation model. Among all published world models~\cite{Hu_2023_Neural_Phd, Hu_2023_gaia1_arXiv, Bogdoll_2023_MUVO_arXiv, Zhang_2023_Copilot4d_ICLR, gaoVistaGeneralizableDriving2024}, MUVO was the only one with code and weights publicly available during time of writing, thus we selected it for our experiments. MUVO is a multimodal world model that uses camera and lidar data and is capable of reconstructing observations in both spaces. Available MUVO weights were trained on a large dataset which was created using the CARLA simulator~\cite{Dosovitskiy17}. It was trained on different driving scenarios in different towns, under different weather conditions, and at different times of the day. The training dataset of MUVO does not contain anomalies and thus establishes the baseline for typical behavior in the context of anomaly detection. Since AnoVox was also generated with CARLA, retraining MUVO for our approach was therefore not necessary.

For image segmentation, all prior works shown in Table~\ref{tab:mask_sota} utilized the Segment Anything Model~\cite{Kirillov_2023_segany_arXiv}. However, SAM was trained in a supervised manner, limiting the use of large-scale, unlabeled datasets as typically available in autonomous driving. Thus, we opted for U2Seg for unsupervised image segmentation~\cite{Niu_2023_Unsupervised_arXiv}. U2Seg is an unsupervised image segmentation model that is capable of generating panoptic segmentation masks by using self-supervised learning and clustering. It would have been beneficial to train U2Seg exclusively on the target domain, but as it was trained on the entirety of ImageNet~\cite{deng_imagenet}, we lacked the necessary resources and used a provided checkpoint.

\textbf{Baseline.} As described in Section~\ref{sec:sota}, there are only two recent anomaly detection models that are trained in a fully unsupervised matter. While both models demonstrate similar performances, \citet{Abati_2019_Latent_IEEECVF} do only provide inference, but not training code for their approach. Thus, we decided to evaluate our approach against MNAD~\cite{Park_2020_Learning_IEEECVF}. MNAD provides code both for prediction and reconstruction tasks but focuses on frame-wise evaluations. We were able to reproduce the experimental results of \citet{Park_2020_Learning_IEEECVF}.

For our evaluation, we trained MNAD on the dataset that was used to train MUVO but sampled each $100^{th}$ frame from it, resulting in 2,725 frames. This ensures that MNAD was trained on images from the same towns, with the same driving conditions, and thus with the same semantic structure as MUVO. The sampling was necessary to prevent overfitting, as MUVO was trained on a much larger dataset. UCSD Ped2, however, which was originally used by \citet{Park_2020_Learning_IEEECVF}, only contains 2,550 images. The data sampling thus allows a dataset size which is comparable to the one used to train MNAD in the experimental setup by \citet{Park_2020_Learning_IEEECVF}. Following~\citet{Park_2020_Learning_IEEECVF}, we trained for 60 epochs. Contrary to our approach, MNAD only localizes anomalies as an intermediate step and uses additional metrics for their final frame-wise score. Based on these intermediate reconstructions, we use the L2 distance to compute pixel-wise anomaly scores.

\section{Evaluation}
\label{sec:eval}

For evaluating UMAD and MNAD, we computed the Average Precision (AP), the False Positive Rate at 95 \% True Positive Rate (FPR$_{95}$), and the Area under the Receiver Operating Characteristic curve (AUROC), as they are common metrics in anomaly detection benchmarks for autonomous driving~\cite{Blum_2019_Fishyscapes_IJCV,Chan_2021_SegmentMeIfYouCan_NIPSTDB}. All results can be found in Table~\ref{tab:eval_results}.

\begin{table}[]
\centering
\resizebox{0.62\textwidth}{!}{%
\begin{tabular}{ccccccccccc}
\rotatebox[origin=c]{70}{$w_{ABS}$} & \rotatebox[origin=c]{70}{$w_{MSE}$} & \rotatebox[origin=c]{70}{$w_{SSIM}$} & \rotatebox[origin=c]{70}{$w_{per}$} & \rotatebox[origin=c]{70}{$w_{temp}$} & \rotatebox[origin=c]{70}{AP $\uparrow$} & \rotatebox[origin=c]{70}{FPR$_{95}$ $\downarrow$} & \rotatebox[origin=c]{70}{AUROC $\uparrow$} & \rotatebox[origin=c]{70}{AP $\uparrow$} & \rotatebox[origin=c]{70}{FPR$_{95}$ $\downarrow$} & \rotatebox[origin=c]{70}{AUROC $\uparrow$} \\ \toprule
 &  &  &  & \multicolumn{1}{l|}{} & \multicolumn{3}{c|}{\textbf{Ground truth}} & \multicolumn{3}{c}{\textbf{SAM}} \\
1 & 0 & 0 & 0 & \multicolumn{1}{l|}{0} &  17.68&  35.56& \multicolumn{1}{l|}{65.23} & 13.72 & 50.58 & 65.16 \\
0 & 1 & 0 & 0 & \multicolumn{1}{l|}{0} &  19.05&  38.92& \multicolumn{1}{l|}{63.61} & 13.77 & 52.22 & 64.93 \\
0 & 0 & 1 & 0 & \multicolumn{1}{l|}{0} &  19.77&  21.26& \multicolumn{1}{l|}{79.03} & 11.43 & 46.79 & 68.26 \\
0 & 0 & 0 & 1 & \multicolumn{1}{l|}{0} &  \textbf{29.90}&  \textbf{16.93}& \multicolumn{1}{l|}{\textbf{83.18}} & \textbf{18.93}& \textbf{42.32}& \textbf{71.88}\\
0 & 0 & 0 & 0 & \multicolumn{1}{l|}{1} &  11.41&  52.70& \multicolumn{1}{l|}{49.15} & 7.11 & 69.26 & 47.72 \\
0 & $\frac{1}{3}$ & $\frac{1}{3}$ & $\frac{1}{3}$ & \multicolumn{1}{l|}{0} &  \underline{27.50}&  \underline{17.81}& \multicolumn{1}{l|}{\underline{82.47}} & \underline{17.11}& 44.01 & 70.83 \\
$\frac{1}{3}$ & 0 & $\frac{1}{3}$ & $\frac{1}{3}$ & \multicolumn{1}{l|}{0} &  26.21&  18.16& \multicolumn{1}{l|}{82.07} & 16.02 & 44.55 & 70.88 \\
$\frac{1}{3}$ & $\frac{1}{3}$ & 0 & $\frac{1}{3}$ & \multicolumn{1}{l|}{0} &  25.52&  20.67& \multicolumn{1}{l|}{79.73} & \underline{17.11}& \underline{43.83}& \underline{71.74}\\
$\frac{1}{3}$ & $\frac{1}{3}$ & $\frac{1}{3}$ & 0 & \multicolumn{1}{l|}{0} &  18.85&  22.88& \multicolumn{1}{l|}{77.73} & 12.85 & 46.44 & 70.20 \\
0 & $\frac{1}{4}$& $\frac{1}{4}$& $\frac{1}{4}$& \multicolumn{1}{l|}{$\frac{1}{4}$} &  25.28&  18.53& \multicolumn{1}{l|}{81.83} & 14.30 & 45.18 & 69.85 \\
$\frac{1}{4}$& 0 & $\frac{1}{4}$& $\frac{1}{4}$& \multicolumn{1}{l|}{$\frac{1}{4}$} &  24.34&  19.39& \multicolumn{1}{l|}{81.27} & 14.74 & 44.87 & 69.95 \\
$\frac{1}{4}$& $\frac{1}{4}$& 0 & $\frac{1}{4}$& \multicolumn{1}{l|}{$\frac{1}{4}$} &  22.28&  21.43& \multicolumn{1}{l|}{79.05} & 16.15 & 45.28 & 70.88 \\
$\frac{1}{4}$& $\frac{1}{4}$& $\frac{1}{4}$& 0& \multicolumn{1}{l|}{$\frac{1}{4}$} &  17.25&  24.15& \multicolumn{1}{l|}{76.92} & 12.60& 48.12& 68.42\\
$\frac{1}{4}$& $\frac{1}{4}$& $\frac{1}{4}$& $\frac{1}{4}$& \multicolumn{1}{l|}{0} &  23.47&  19.04& \multicolumn{1}{l|}{81.34} & 15.55 & 44.35 & 71.12 \\
$\frac{1}{5}$& $\frac{1}{5}$& $\frac{1}{5}$& $\frac{1}{5}$& \multicolumn{1}{l|}{$\frac{1}{5}$} &  22.38&  19.71& \multicolumn{1}{l|}{80.74} & 14.52 & 45.02 & 70.21 \\ \midrule
 &  &  &  & \multicolumn{1}{l|}{} & \multicolumn{3}{c|}{\textbf{U2Seg}} & \multicolumn{3}{c}{\textbf{Max. Value}} \\
1 & 0 & 0 & 0 & \multicolumn{1}{l|}{0} &  14.04&  60.20& \multicolumn{1}{l|}{59.55} &  19.00&  59.68&  40.68\\
0 & 1 & 0 & 0 & \multicolumn{1}{l|}{0} &  14.54&  60.98& \multicolumn{1}{l|}{59.93} &  18.86&  59.52&  40.76\\
0 & 0 & 1 & 0 & \multicolumn{1}{l|}{0} &  12.17&  58.44& \multicolumn{1}{l|}{62.88} &  10.87&  67.03&  33.30\\
0 & 0 & 0 & 1 & \multicolumn{1}{l|}{0} &  \textbf{18.88}&  56.74& \multicolumn{1}{l|}{64.77} &  17.26&  57.68&  42.55\\
0& 0 & 0 & 0 & \multicolumn{1}{l|}{1} &  9.02&  68.89& \multicolumn{1}{l|}{54.44} &  11.01&  74.23&  25.97\\
0 & $\frac{1}{3}$ & $\frac{1}{3}$ & $\frac{1}{3}$ & \multicolumn{1}{l|}{0} &  17.70&  56.76& \multicolumn{1}{l|}{65.09} &  \underline{20.97}&  \underline{52.01}&  \textbf{48.57}\\
$\frac{1}{3}$ & 0 & $\frac{1}{3}$ & $\frac{1}{3}$ & \multicolumn{1}{l|}{0} &  17.13&  \underline{56.73}& \multicolumn{1}{l|}{\textbf{65.50}} &  18.71&  52.63&  47.85\\
$\frac{1}{3}$ & $\frac{1}{3}$ & 0 & $\frac{1}{3}$ & \multicolumn{1}{l|}{0} &  \underline{17.99}&  57.07& \multicolumn{1}{l|}{\underline{65.47}} &  \textbf{21.91}&  \textbf{51.83}&  \underline{48.49}\\
$\frac{1}{3}$ & $\frac{1}{3}$ & $\frac{1}{3}$ & 0 & \multicolumn{1}{l|}{0} &  13.64&  58.31& \multicolumn{1}{l|}{63.97} &  18.51&  60.24&  40.52\\
0 & $\frac{1}{4}$& $\frac{1}{4}$& $\frac{1}{4}$& \multicolumn{1}{l|}{$\frac{1}{4}$} &  17.08&  \textbf{56.54}& \multicolumn{1}{l|}{65.13} &  16.69&  56.77&  43.76\\
$\frac{1}{4}$& 0 & $\frac{1}{4}$& $\frac{1}{4}$& \multicolumn{1}{l|}{$\frac{1}{4}$} &  16.35&  56.92& \multicolumn{1}{l|}{65.18} &  15.44&  57.88&  42.90\\
$\frac{1}{4}$& $\frac{1}{4}$& 0 & $\frac{1}{4}$& \multicolumn{1}{l|}{$\frac{1}{4}$} &  17.15&  57.09& \multicolumn{1}{l|}{64.97} &  18.44&  56.67&  43.68\\
$\frac{1}{4}$& $\frac{1}{4}$& $\frac{1}{4}$& 0& \multicolumn{1}{l|}{$\frac{1}{4}$} &  12.17&  58.44& \multicolumn{1}{l|}{62.88} & 16.05& 63.27& 37.25\\
$\frac{1}{4}$& $\frac{1}{4}$& $\frac{1}{4}$& $\frac{1}{4}$& \multicolumn{1}{l|}{0} &  17.16&  56.84& \multicolumn{1}{l|}{62.88} &  19.86&  53.36&  47.22\\
$\frac{1}{5}$& $\frac{1}{5}$& $\frac{1}{5}$& $\frac{1}{5}$& \multicolumn{1}{l|}{$\frac{1}{5}$} &  16.38&  57.06& \multicolumn{1}{l|}{65.04} &  20.01&  56.84&  43.82\\ \midrule

\rotatebox[origin=c]{70}{$w_{ABS}$} & \rotatebox[origin=c]{70}{$w_{MSE}$} & \rotatebox[origin=c]{70}{$w_{SSIM}$} & \rotatebox[origin=c]{70}{$w_{per}$} & \rotatebox[origin=c]{70}{$w_{temp}$} & \rotatebox[origin=c]{70}{AP $\uparrow$} & \rotatebox[origin=c]{70}{FPR$_{95}$ $\downarrow$} & \rotatebox[origin=c]{70}{AUROC $\uparrow$} & \rotatebox[origin=c]{70}{AP $\uparrow$} & \rotatebox[origin=c]{70}{FPR$_{95}$ $\downarrow$} & \rotatebox[origin=c]{70}{AUROC $\uparrow$} \\ \toprule

 &  &  &  & \multicolumn{1}{l|}{} & \multicolumn{3}{c|}{\textbf{No Mask}} & \multicolumn{3}{c}{\textbf{Single Mask}} \\
1 & 0 & 0 & 0 & \multicolumn{1}{l|}{0} &  6.80&  78.19& \multicolumn{1}{l|}{60.19} &  5.04&  93.57&  50.54\\
0 & 1 & 0 & 0 & \multicolumn{1}{l|}{0} &  7.03&  78.49& \multicolumn{1}{l|}{60.68} &  5.04&  93.57&  50.53\\
0 & 0 & 1 & 0 & \multicolumn{1}{l|}{0} &  4.72&  50.87& \multicolumn{1}{l|}{73.02} &  5.83&  92.83&  51.12\\
0 & 0 & 0 & 1 & \multicolumn{1}{l|}{0} &  \textbf{10.86}&  \textbf{32.91}& \multicolumn{1}{l|}{\textbf{79.51}} &  \underline{10.40}&  88.49&  53.26\\
0 & 0 & 0 & 0 & \multicolumn{1}{l|}{1} &  4.09&  73.37& \multicolumn{1}{l|}{53.05} &  5.06&  93.57&  50.59\\
0 & $\frac{1}{3}$ & $\frac{1}{3}$ & $\frac{1}{3}$ & \multicolumn{1}{l|}{0} &  \underline{9.51}&  \underline{37.37}& \multicolumn{1}{l|}{53.05} &  \textbf{12.66}&  \textbf{86.31}&  \textbf{54.48}\\
$\frac{1}{3}$ & 0 & $\frac{1}{3}$ & $\frac{1}{3}$ & \multicolumn{1}{l|}{0} &  9.29&  38.99& \multicolumn{1}{l|}{\underline{78.70}} &  8.88&  89.93&  52.59\\
$\frac{1}{3}$ & $\frac{1}{3}$ & 0 & $\frac{1}{3}$ & \multicolumn{1}{l|}{0} &  9.42&  42.34& \multicolumn{1}{l|}{76.24} &  8.83&  89.94&  52.54\\
$\frac{1}{3}$ & $\frac{1}{3}$ & $\frac{1}{3}$ & 0 & \multicolumn{1}{l|}{0} &  6.93&  52.40& \multicolumn{1}{l|}{72.32} &  5.05&  93.56&  50.64\\
0 & $\frac{1}{4}$& $\frac{1}{4}$& $\frac{1}{4}$& \multicolumn{1}{l|}{$\frac{1}{4}$} &  8.29&  39.37& \multicolumn{1}{l|}{77.51} &  8.88&  89.93&  52.57\\
$\frac{1}{4}$& 0 & $\frac{1}{4}$& $\frac{1}{4}$& \multicolumn{1}{l|}{$\frac{1}{4}$} &  8.14&  40.26& \multicolumn{1}{l|}{77.17} &  10.37&  \underline{88.48}&  \underline{53.35}\\
$\frac{1}{4}$& $\frac{1}{4}$& 0 & $\frac{1}{4}$& \multicolumn{1}{l|}{$\frac{1}{4}$} &  8.50&  44.02& \multicolumn{1}{l|}{75.05} &  7.30&  91.39&  51.79\\
$\frac{1}{4}$& $\frac{1}{4}$& $\frac{1}{4}$& 0& \multicolumn{1}{l|}{$\frac{1}{4}$} &  6.16&  53.28& \multicolumn{1}{l|}{71.14} & 4.30& 94.29& 50.26\\
$\frac{1}{4}$& $\frac{1}{4}$& $\frac{1}{4}$& $\frac{1}{4}$& \multicolumn{1}{l|}{0} &  8.83&  40.07& \multicolumn{1}{l|}{77.62} &  8.83&  89.93&  52.57\\
$\frac{1}{5}$& $\frac{1}{5}$& $\frac{1}{5}$& $\frac{1}{5}$& \multicolumn{1}{l|}{$\frac{1}{5}$} &  8.11&  41.12& \multicolumn{1}{l|}{76.69} &  8.07&  90.66&  52.18\\ \midrule
\multicolumn{5}{c|}{\textbf{MNAD}~\cite{Park_2020_Learning_IEEECVF}} & 6.37 & 89.61 & \multicolumn{1}{l|}{61.96} & --- & --- & --- \\ \bottomrule
\vspace{2pt}
\end{tabular}
}
\caption{\textbf{Evaluation Results.} We show evaluation results for six settings of our model and MNAD with \textbf{best} and \underline{second-best} results highlighted. The experimental results for a setup comprising a ground truth image segmentation map, segmentation maps generated with U2Seg and SAM, a setup where the instance-wise maximum anomaly score is selected instead of the average anomaly score, a setup without an image segmentation map, and a setup where only the instance with the highest anomaly score is left in the anomaly map are depicted. All evaluation metrics in $\%$.}
    \label{tab:eval_results}
\end{table}

\textbf{Ablation Studies.} In order to better understand our method, we perform a set of ablation studies. First, next to utilizing U2Seg, we were interested in the possible performance gains of using SAM~\cite{Kirillov_2023_segany_arXiv} or ground truth masks. SAM is a zero-shot image segmentation model that is also used by SotA anomaly detection models. While SAM was trained with labeled data, it performs well in the context of zero-shot inference. On the other hand, we also wanted to understand the effects of not refining our anomaly map with masks at all.

Second, rather than averaging all anomaly scores per mask, we were interested in whether picking the maximum value, inspired by~\citet{liuHybridVideoAnomaly2021}, impacts the performance. On a similar note, we were also interested in picking only the mask with the highest anomaly score, neglecting the rest.

\textbf{Experimental Results.} Here, we present our findings on the performance of UMAD compared to the MNAD baseline, as well as our ablation studies. Since the visual differences and the temporal difference are normalized, they can be individually weighted and combined in order to form an anomaly map. This process is done in the Weighted Fusion Model. In the following, we also evaluate the impact of the single difference metrics and their combinations.

Since MNAD does not use masks, we first compare the pixel-wise L2 distance of MNAD to the similarly calculated squared error of our model on the raw pixel-wise output without masks. The experimental results indicate that utilizing world models instead of autoencoders is beneficial for anomaly detection in autonomous driving: With UMAD in the setting of using the \textit{squared error} as visual difference, the AP is $7.03 \%$ and the FPR$_{95}$ $78.49 \%$. The AP is thus $10.36 \%$ higher and the FPR$_{95}$ $12.41 \%$ lower than in the evaluation of MNAD. Even better results can be achieved when using the \textit{perceptual difference} for visual difference: Here, UMAD achieves by far the highest average precision, lowest FPR$_{95}$, and highest AUROC in the pixel-wise setup without masks: AP is $70.49 \%$ higher and FPR$_{95}$ $63.27 \%$ lower than in the experimental results for MNAD. When using masks that are generated with the unsupervised image segmentation model U2Seg and the perceptual difference as visual difference, it is possible to achieve an AP that is $196.39 \%$ higher than the AP in the evaluation of MNAD, indicating that using masks for anomaly detection is highly beneficial. Using a combination of the mean squared error, the SSIM, the perceptual difference and the temporal difference, the FPR$_{95}$ is $36.90 \%$ lower than in the evaluation of MNAD.

When using the zero-shot image segmentation approach SAM, which is also used as an image segmentation approach in prior anomaly detection models, it is possible to furthermore improve our experimental results. With the perceptual difference as visual difference in the setup, AP in this setup is $18.93 \%$, FPR$_{95}$ is $52.77 \%$ smaller and AUROC $16.01 \%$ higher than in the respective results for MNAD. To evaluate the full potential of utilizing masks for anomaly detection, we furthermore evaluated UMAD with masks from a ground truth instance segmentation map. This setup achieved by far the best experimental results, again showing the huge potential of leveraging masks in anomaly detection: The best AP score with this experimental setup is $29.90 \%$, the best FPR$_{95}$ is $16.93 \%$, and the best AUROC is $83.18 \%$.

In the prior experimental setups, the average anomaly score of the masks was used for evaluation. Interestingly, we found that the perceptual difference is not suitable for anomaly detection when assigning the maximum anomaly score to masks rather than their average score. Generally, however, substituting the average anomaly score per instance with the maximum score, did not achieve better results. Worst results are achieved when only considering the instance with the highest anomaly score. We found that often not the anomalous object, but a different object in the observation has the highest anomaly score. This then results in completely ignoring the abnormal object.

\section{Conclusion}
\label{sec:conclusion}

    In this work, we presented UMAD, the first fully unsupervised anomaly detection model for autonomous driving which utilizes generative world models and is capable of combining anomaly scores and image segmentation approaches for masked anomaly detection. We find that utilizing world models in combination with image segmentation approaches is highly beneficial for anomaly detection in autonomous driving. Furthermore, we demonstrate that perceptual difference, compared to other approaches, is highly suitable for generating reconstruction errors in generative anomaly detection models. UMAD sets a new baseline in unsupervised anomaly detection for autonomous driving by achieving a FPR$_{95}$ reduction of $36.90 \%$ on the challenging AnoVox benchmark.

    \textbf{Limitations and Outlook.} Since the reconstruction quality of the MUVO world model was, under some circumstances, highly fluctuating and affecting the anomaly detection performance, we are interested in evaluating a more recent world model~\cite{gaoVistaGeneralizableDriving2024} in the future. Also, since we lacked the computational resources to train U2Seg on our target domain, a domain shift exists. Once unsupervised segmentation models become less compute-intensive, we are interested in ablating the effects of such a domain shift. Furthermore, the perceptual difference utilizes weights which are pre-trained on the ImageNet \cite{deng_imagenet} dataset. Despite achieving promising results using the perceptual difference as visual difference, we are interested in evaluating whether further improvements can be achieved when the VGG network for the perceptual difference is trained on the same dataset which is used to train the underlying world model. Finally, we only performed experiments in a simulated environment. In the future, we want to apply the approach to real-world driving scenarios.

\section*{Acknowledgment}
\label{sec:ackno}

This work results from the just better DATA project supported by the German Federal Ministry for Economic Affairs and Climate Action (BMWK), grant number 19A23003H.

\newpage
\bibliography{egbib}
\end{document}